\documentclass[preprint,12pt]{elsarticle}

\usepackage{amssymb,amsmath}
\usepackage{epstopdf}

\journal{ar$\chi$iv}

\begin{document}

\begin{frontmatter}



\title{Spike Event Based Learning in Neural Networks}


\author{J. A. Henderson, T. A. Gibson, J. Wiles}

\address{}

\begin{abstract}

A scheme is derived for learning connectivity in spiking neural networks. The scheme learns instantaneous firing rates that are conditional on the activity in other parts of the network. The scheme is independent of the choice of neuron dynamics or activation function, and network architecture. It involves two simple, online, local learning rules that are applied only in response to occurrences of spike events. This scheme provides a direct method for transferring ideas between the fields of deep learning and computational neuroscience. This learning scheme is demonstrated using a layered feedforward spiking neural network trained self-supervised on a prediction and classification task for moving MNIST images collected using a Dynamic Vision Sensor.

\end{abstract}


\end{frontmatter}



\section{Introduction}
Methods in deep learning for training neural networks (NNs) have been very successfully applied to a range of datasets, performing tasks at levels approaching human performance, such as image classification \citep{ILSVRCarxiv14}, object detection \citep{ILSVRCarxiv14} and speech recognition \citep{ASLPIEEE2012, ICASSP2013, SPM2012}. Along with these experimental successes, the field of deep learning is rapidly developing theoretical frameworks in representation learning \citep{PAMIIEEE2013, bengio2014deep, NN:2014} including understanding the benefits of different types of non-linearities in neuron activation functions \citep{ICML2010}, disentanglement of inputs by projecting onto hidden layer manifolds,  model averaging with techniques like maxout and dropout \citep{ICML2013, JMLR2014} and assisting generalization through corruption of input with denoising autoencoders \citep{JMLR:2010}.

These types of experimental and theoretical work are necessary to effectively build and understand systems like brains that are capable of learning to solve real world problems. Many of the successes of deep learning are a result of a broad inspiration from biology; however, there is a large gap in understanding how the principles of deep learning are related to those of the brain. Some elements of deep learning may well inspire discoveries in brain function. Equally, deep learning systems are still inferior to the brain in aspects such as memory, thus efforts to develop models that bridge between deep learning and neuroscience are likely to be mutually beneficial.

The neuron models commonly used in deep learning are abstracted away from neuron models that are used in computational neuroscience to model biological neurons. Spiking is a salient feature of biological neurons that is not typically present in deep learning networks. It is not yet understood why the brain uses spiking dynamics; for the purposes of machine learning it would be useful to know what if any advantages spiking dynamics confers spike based NN learning algorithms over other types of NN learning algorithms, rather than advantages that are otherwise useful in implementing algorithms in biology such as energy efficiency and robustness. Dynamical systems like spiking networks appear more naturally suited to processing continuous time temporal data than state machines, as deep networks are usually implemented, but this idea is yet to be demonstrated experimentally on machine learning tasks.

In an effort to bridge this gap in understanding between spike, and non-spike based NN learning systems, and develop systems for processing event based, continuous time data, this paper develops a scheme for learning connectivity in a spiking neural network (SNN). The scheme is based upon learning conditional instantaneous firing rates, linking it to many of the statistical frameworks previously developed in deep learning that are based on conditional probabilities. However, our scheme is fundamentally different to most methods used in deep learning as the learning rules are based solely on the activity of the neurons in the network and are the same, independent of the choice of neuron dynamics or activation function unlike gradient descent methods \citep{NAT:1986}, and they can be implemented online and do not require periods of statistical sampling from the model unlike energy based methods \citep{NC:2006}. In addition, the learning scheme is local, meaning that modifying a connection only requires knowledge of the activity of the neurons it connects, not neurons from a distant part of the network, unlike gradient descent and energy based methods \citep{NAT:1986,NC:2006}.  From a perspective of biological plausibility, this means neurons do not have to make assumptions about, or communicate to each other their dynamics or activation function and associated parameters in order to correctly learn, and the system can be run online without interrupting processing with periods of sampling for learning.

This paper describes a general scheme for event based learning in SNNs. This scheme is demonstrated on a network similar to that commonly used in deep learning, specifically, a feedforward layered network architecture with rectified linear units, and piecewise constant temporal connectivity. Dropout is utilized to show that many ideas from deep learning can be directly imported into SNNs using this learning scheme. An event based dataset of moving MNIST digits collected using an DVS camera \citep{DVSMNIST, MNIST,IEEESSC:2013} is used to train the network for both prediction and classification tasks. 

\section{Learning Theory}
\label{SecSONL}
We begin by developing a method for learning the connectivity of a supervised output neuron. The discussion will be framed in reference to learning in a network operating continuously in time with temporally delayed connectivity and temporally encoded input signals since spiking neurons are usually modelled as dynamical systems; however, the results are also applicable to networks operating in discrete timesteps (as is usual in implementations of SNNs with current standard computer architecture), with or without temporally delayed connectivity and temporally encoded inputs, so they can also be applied to traditional artificial neural networks performing static image classification, for example.

\begin{figure}
\includegraphics[width=1\textwidth]{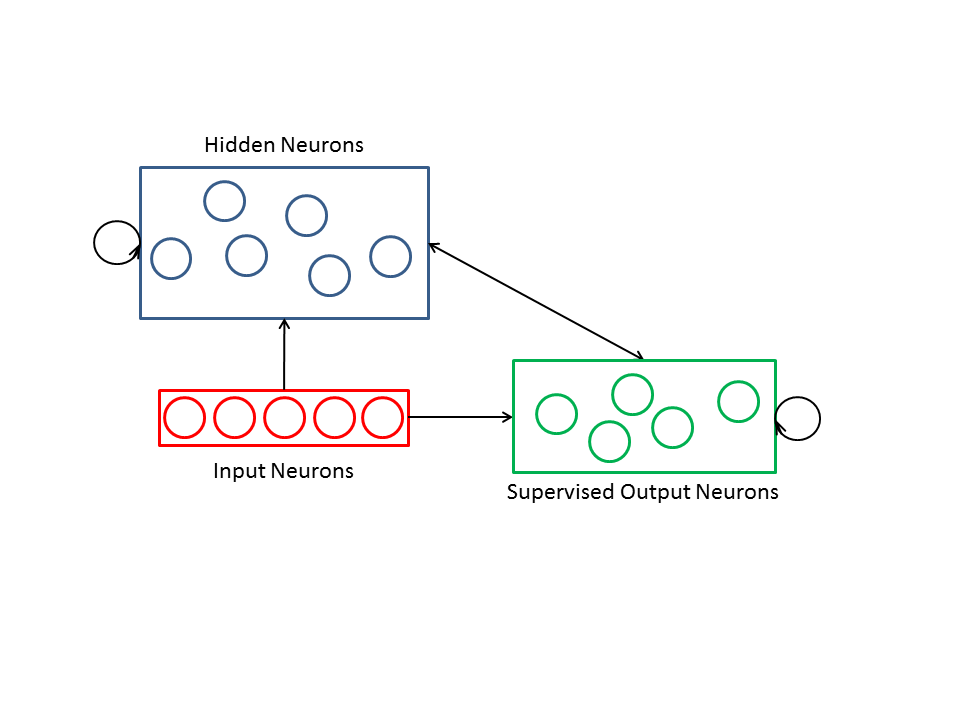}
\caption{\label{FIGNetDiag} Schematic of a general neural network consisting of input neurons controlled by an external source, hidden neurons and supervised output neurons. Connectivity between neurons is unrestricted, directed connections are allowed between every pair of neurons including self connections, as are temporally delayed connections.}
\end{figure}

Figure \ref{FIGNetDiag} shows a general network containing input neurons whose activity is determined by an external source, hidden neurons, and supervised output neurons. At present we do not assume any particular connection architecture, nor do we specify the dynamics or activation functions of the individual neurons. 

We first consider learning the input, $Q_o(t)$, that a supervised output neuron $o$ receives from the network at time $t$. We need to identify the mathematical quantity that $o$ should learn - the quantity that $Q_o(t)$ should approximate. Note that $Q_o(t)$ may be calculated from two sets of quantities only. The first quantities are the weights $W$, that connect the activity of the network to $o$, potentially including self connections and connections that have a temporal delay. We call these connections {\it weights} as this is the terminology usually used in machine learning; however, these connections can also be imagined as propagator functions whose value varies with temporal delay. The second quantity is the history of activity of the neurons in the network, $H$, which since we are primarily interested in SNNs and event based learning, we model as a set of one dimensional Dirac delta functions in time that may be normalized to non unit integral to allow a neuron's spike strength to be a real value, instead of only binary as in many SNN models. However, since it allows the use of more intuitive terminology of spike rates, instead of spike strength rates, we assume that a spike with real valued strength is equivalent to a sum of simultaneous unit strength spikes and possibly one partial unit spike. Alternatively, this equivalence holds if spike strengths are restricted to a unit value and simultaneous spikes are not allowed. In any case the mathematical description and quantitative results are unchanged aside from a possible conversion function if simultaneous spikes are not considered to combine additively into a single real valued spike and vice-versa.

Assuming $W$ is fixed after learning, the only time varying quantity that can be used in calculating $Q_o(t)$ is $H(t)$, so we re-parametrize $Q_o(t)$ to $Q_o(H)$. We then propose that a sensible output of the network to $o$ is $Q(o|H)$, the mean conditional instantaneous spike rate (during training) of supervised output neuron $o$, given activity $H$ in the network. Integrating $Q$ over a time period gives an expected number of spikes. Thus, in the case of a network operating in discretized time, as is common in the implementation of artificial SNNs in code,  $Q(o|H)$ can be interpreted as an expected number of spikes $o$ given $H$, where the integral over a small discrete timestep is understood.  That is, if we observe activity $H$ in the network $n$ times during training, and $o$ spikes $n_o$ times during timesteps coinciding with those $n$ occurrences of $H$, then after training if we observe $H$ again, the output that should have been learnt and produced by the network at that timestep is $n_o/n$. If only single unit strength spikes are allowed at each timestep, then this output can be interpreted as $P(o|H)$, the probability of $o$ spiking during the given timestep, given $H$. This interpretation is important in connecting the focus on probability distributions in machine learning with the focus on spike and rate coded networks in computational neuroscience \citep{NRN:2010}. 

Of course if $H$ includes the full history of the network's activity from inception, then only one sample trajectory of $H$ will be observed and used for learning. However, we assume that $W$ approaches zero as the connection time delay becomes large, meaning that only some recent history of activity in the network is used in calculating $Q(o|H)$. Thus, a variety of different $H$ will be observed during training. In any practical network $W$ will be finitely parametrized, so for any particular $H_k$ the parameters modified in learning $Q(o|H_k)$ will in effect learn for both a range of other $H$ that are slight perturbations of $H_k$ and also use the same parameters, as well as very different $H$ that only share a portion of the same parameters. 

If we assume the neurons in the network are spiking neurons, then there are only four events that occur within the network at which to apply learning rules that modify $Q_o(H)$. They are (i) when an input neuron spikes, (ii) when a hidden neuron spikes, (iii) when a supervised output neuron spikes due to supervision, and (iv) when a supervised output neuron spikes due to its own dynamics or activation function. Modification could also be made continuously at all times, at randomly generated time points, or according to a temporally periodic function; however, we proceed concentrating on the spiking events.

\begin{figure}
\includegraphics[width=1\textwidth]{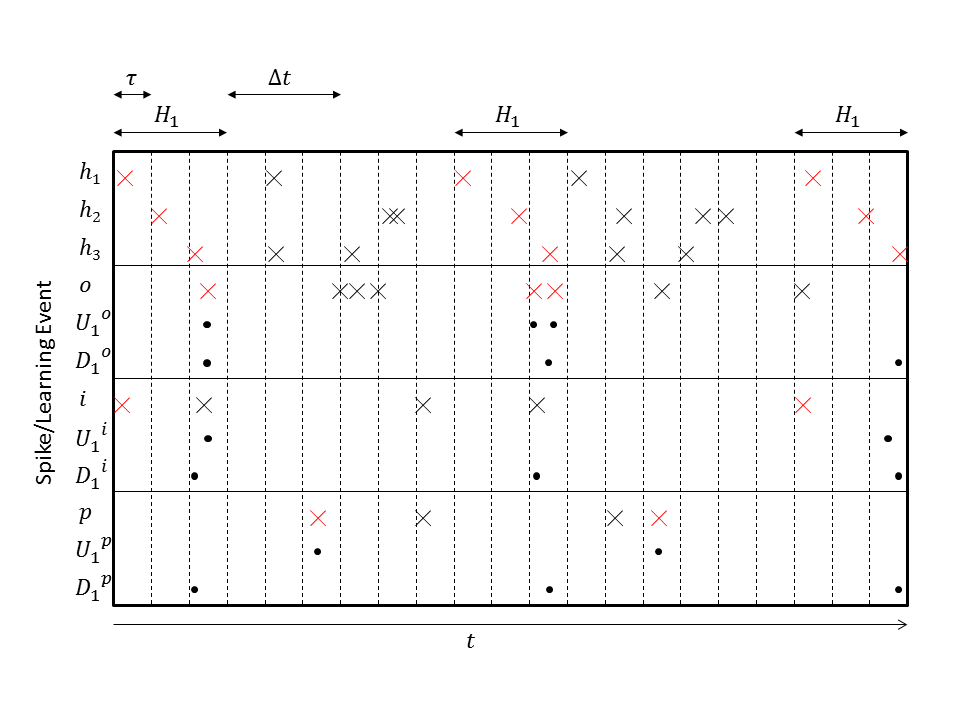}
\caption{ \label{FIGSpikeSchem}Schematic showing the timing of events in the operation and learning of the network.  Neuron spike events are indicated by crosses, learning rule applications are indicated by dots. Vertical dotted lines indicate timesteps of width $\tau$ used in the network's implementation in code (the network can, given suitable hardware, be operated as a dynamical system in continuous time). Note that although variation in the location of spikes within a timestep occurs, this variation is not resolved by a time stepped implementation, additionally single neurons can spike multiple times within a single timestep. This diagram focuses upon learning for a particular hidden neuron activity pattern $H_1$ indicated by red crosses. (i) For output neuron $o$, $U^o_1$ is applied each time $o$ spikes in conjunction with $H_1$, $D^o_1$ is applied each time $H_1$ is observed. (ii) For input neuron $i$, each time $i$ spikes in conjunction with the beginning of $H_1$, $U^i_1$ is applied at the conclusion of $H_1$ as $H_1$ must be observed in order to identify the connections to modify. $D^i_1$ is applied each time $H^i_1$ occurs. (iii) For prediction neuron $p$, each time $p$ spikes time $\Delta t$ after $H_1$ occurs, $U^p_1$ is applied. $D^p_1$ is applied each time $H_1$ occurs. }
\end{figure}

Let $D$ be a function that is applied to $Q_o(H)$ when $H$ occurs (a combination of events (ii)), and let $U$ be a function that is applied to $Q_o(H)$ for each supervised spike $o$ (an event (i)) that co-occurs with $H$ (see Fig. \ref{FIGSpikeSchem}). After some period of training time $t$, we have a series of iterated applications of $D$ and $U$ applied to the initial value $Q^{0}_o(H)$
\begin{equation}
\label{QT}
Q^{t}_o(H)=\left (D \circ U \circ ... \circ U \right) \circ ...\circ \left( D \circ U \circ ... \circ U \right) \circ Q^{0}_o(H),
\end{equation}
where the brackets group operations for each occurrence of $H$. 

To find a relation between $U$ and $D$, suppose now that the initial value $Q^{0}_o(H)=Q(o|H)$ as we desire. Clearly, we also require that $Q^{t}_o(H)\approx Q(o|H)$, meaning that the application of the learning rules $D$ and $U$ do not cause the output to significantly deviate from the desired value. Note that we cannot require strict equality due to the stochastic nature of the event occurrences. If we choose 
\begin{equation}
\label{EQUDJ}
U=D^{J(Q^{t'}_o(H))},
\end{equation}
where superscripts indicate composition, not exponentiation and $J$ is an unknown function still to be determined, then with the initial value $Q^{0}_o(H)=Q(o|H)$,  Eq. (\ref{QT}) becomes
\begin{equation}
Q^{t}_o(H)=\left (D \circ D^{J(Q_o^{t''}(H))}\circ ... \circ D^{J(Q_o^{t''}(H))} \right) \circ ...\circ \left( D \circ D^{J(Q_o^{t'}(H))} \circ ... \circ D^{J(Q_o^{t'}(H))} \right) Q(o|H).
\end{equation}

Let $N$ be the number of occurrences of $H$. For $Q^t_o(H)\approx Q(o|H)$, we require that all the applications of $D$ and $U$ approximately cancel, i.e.
\begin{equation}
N+\sum J(Q^{t'}_o(H))\approx 0.
\end{equation}
The expected number of applications of $U$ is $N Q(o|H)$, and we require that $Q^t_o(H)\approx Q(o|H)$ at all points in this sequence of applications of $D$ and $U$, so we have
\begin{equation}
 J(Q_o(H)) N Q(o|H) \approx-N.
\end{equation}
However, since we do not know $Q(o|H)$ {\it a priori} we use the network's current estimate $Q_o(H)$ instead and set
\begin{equation}
 J(Q_o(H))=- \frac{1}{Q_o(H)}.
\end{equation}

Using Eq. (\ref{EQUDJ}) we now have the following relation between the function $D$ that is applied when $H$ occurs, and the function $U$ that is applied when $o$ spikes due to supervision
\begin{equation}
\label{UD}
U=D^{-\frac{1}{Q_o}}.
\end{equation}
This requires that $D$ has a unique inverse, and $D^{-1}$ can be generalized in such a way as to be applied a fractional number of times.

In the above we required that $Q^t_o(H)\approx Q(o|H)$ at all points in a sequence of applications of $D$ and the $U$. This implies that any single application of either $D$ or $U$ when $Q_o(H)\approx Q(o|H)$, can only change $Q_o(H)$ by a small (but not necessarily fixed) amount 
\begin{equation}
\label{resU}
Q_o-\epsilon_U\le U(Q_o)\le Q_o +\epsilon_U,
\end{equation}
\begin{equation}
\label{resD}
Q_o-\epsilon_D\le D(Q_o)\le Q_o +\epsilon_D,
\end{equation}
which using (\ref{UD}) leads to the relation
\begin{equation}
\label{releUeD}
\epsilon_D=Q_o\epsilon_U.
\end{equation}

The required range of $Q_o$ is $[0,\infty)$. To ensure that $\epsilon_U$ remains small as $Q_o\rightarrow 0$, we require $\epsilon_D$ be chosen so that in the $\lim_{Q \to 0}$, $\frac{\epsilon_D}{Q}$ remains finite. Alternatively it would be possible to insert noise spikes, for example Poisson noise with rate $m$ into the supervision to fix a minimum target value of $Q_o$ to $m$, hence bounding $Q_o>0$ and eliminating the divergence in Eq. (\ref{releUeD}). After learning this noise can be stopped and subtracted from the learnt value of $Q(o|H)$. In most cases the maximum value of $Q$ will be finite, and hence $\epsilon_D$ and $\epsilon_U$ can be chosen to give sufficiently small changes.

To avoid $Q_o$ converging to an unwanted value, this learning scheme must have only a single globally stable fixed point $Q_o=Q(o|H)$. This means that $U(Q)$ and $D(Q)$ cannot both have fixed points at any $Q$. We therefore adjust Eqs.  (\ref{resU}) and (\ref{resD}) to
\begin{eqnarray}
\label{resU2}
Q_o-\epsilon_U \le U(Q_o)<Q_o & \rm{or} & Q_o<U(Q_o) \le Q_o +\epsilon_U,
\end{eqnarray}
and
\begin{eqnarray}
\label{resD2}
Q_o<D(Q_o) \le Q_o +\epsilon_D & \rm{or}&  Q_o-\epsilon_D \le D(Q_o)<Q_o.
\end{eqnarray}

We choose between either the two left, or two right options in (\ref{resU2}) and (\ref{resD2}) by considering the stability of the fixed point $Q(o|H)$ for each of these choices. Taking equalities in the above equations and using Eq. (\ref{releUeD}), the total change to $Q_o$ after $N$ applications of $D$ and an expected $Q(o|H)N$ applications of $U$ is
\begin{equation}
\Delta \approx \pm NQ_o\epsilon_U \mp Q(o|H)N\epsilon_U.
\end{equation}
If $Q_o>Q(o|H)$ we require $\Delta<0$, and if $Q_o<Q(o|H)$ we require $\Delta>0$. This implies the following choice for our learning rule restrictions
\begin{equation}
\label{LRD}
Q_o-\epsilon_D\le D(Q_o)<Q_o,
\end{equation}
\begin{equation}
\label{LRU}
Q_o<U(Q_o) \le  Q_o +\epsilon_U,
\end{equation}
that is, $D$ slightly decreases $Q_o$ and $U$ slightly increases $Q_o$. 

\section{Application to Learning Layers of Autoencoders}

\begin{figure}
\includegraphics[width=1\textwidth]{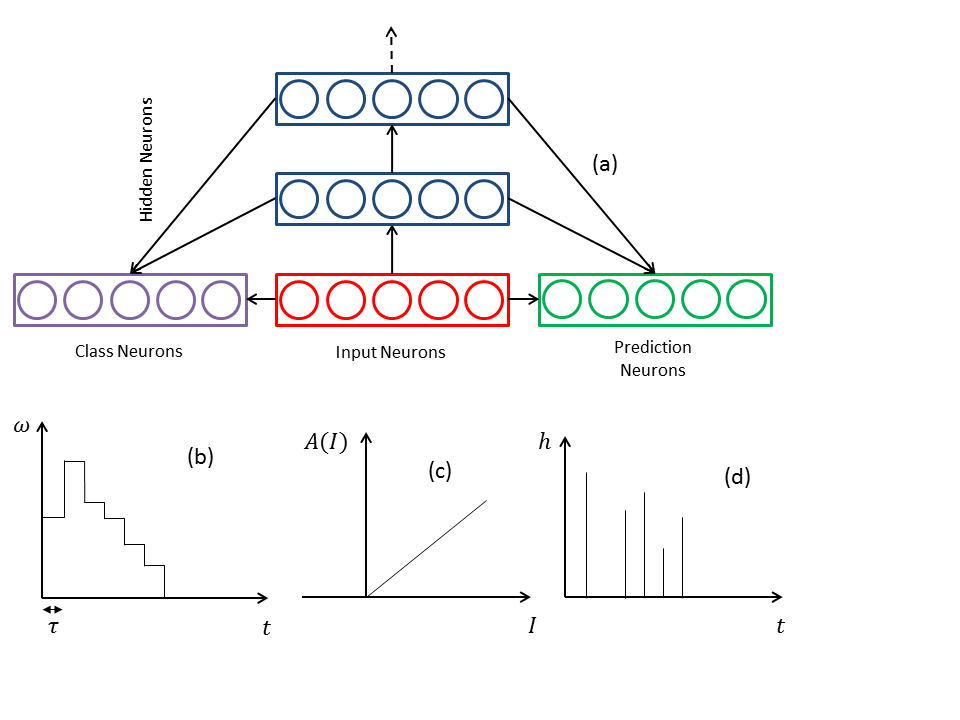}
\caption{\label{FIGLayNetDiag} (a) Architecture of the feedforward layered network. (b) Illustration of piecewise constant connectivity between two neurons in Eq. (\ref{EQwi}). (c) Rectified linear unit activation function for hidden neurons used in this network. (d) Illustration of the spiking activity of a hidden neuron, vertical lines indicate the presence of a Dirac delta function, with height corresponding to different normalizations of each individual Dirac delta function.}
\end{figure}

We now outline a demonstration of this learning scheme. A standard method for training an unsupervised deep feedforward network is to train each pair of layers successively as autoencoders \citep{bengio2014deep} so that each layer encodes the activity of the layer below it, see Fig. \ref{FIGLayNetDiag}. The learning rules described in Sec. \ref{SecSONL} can be used to learn layers of autoencoders by replacing the supervised output neuron $o$ , with an input neuron $i$ that self-supervises, and by reversing the direction of connectivity so that $i$ learns to output $Q(i|H)$, where $H$ is now the future activity of the hidden neurons in the layer above $i$, since causality is reversed from the previous case; the input layer causes activity in the hidden layer above, see Fig. \ref{FIGSpikeSchem}.

Using this method, the hidden layers learn so that by observing a period of hidden layer activity, the activity of the layer below at the beginning of that observation can be inferred. The activity of the hidden layer and the connectivity between layers acts like, and encodes a short term memory.

In this demonstration we also include a layer of prediction neurons that predict the activity of the input layer at a specified time period in the future. These neurons are supervised by the activity of the input layer with the corresponding prediction time period delay, see Fig. \ref{FIGSpikeSchem}. We also include a layer of digit classification neurons that are trained as for a supervised output neuron $o$, see Fig. \ref{FIGSpikeSchem}.

So far we have not needed to specify the dynamics or activation function of the hidden neurons in the network in order to develop this learning scheme. Spiking neuron models in computational neuroscience are often dynamical systems modeled using differential equations \citep{izhikevichbook}. In contrast, neurons in machine learning are typically characterized by an activation function of the neuron's input \citep{bengio2014deep}. Any of these types of neuron models could be employed here; however, we choose rectified linear units (ReLUs) that are commonly used in deep learning networks \citep{bengio2014deep}. The form we use here is 
\begin{align}
A(I)&=I, &I>0, \nonumber \\
&= 0, &I \leq 0,
\end{align}
see Fig. \ref{FIGLayNetDiag}c.

\subsection{Weight Update Rules}

We have so far developed rules for learning a value $Q_o$ to approximate $Q(o|H)$; however, we have not yet discussed rules for modifying $W$ that are necessary for implementation in a network. Before these rules can be determined, the formula for calculating $Q_o$ from $H$ and $W$ needs to be chosen. The most common choice is to use the product of $H$ and $W$ summed across all neurons in the layer below and integrated across time in the case of time delay connections. We use this same choice here
\begin{equation}
\label{EQQdef}
Q_o(t) =\sum_j \int_{0}^t h_j(t')w_j(t-t')dt',
\end{equation}
where $h_j$ is a hidden neuron connected to $o$ and $w_j$ is the corresponding connection between them. Other choices are possible and may have advantages over this choice, though this is left for future investigation. We also need to choose a parametrization for $W$. A wide variety of choices could be made here such as sums of continuous functions, or convolution kernels acting across different sets of $j$, as is done in convolutional neural networks by modifying  (\ref{EQQdef}) to include a convolution across $j$ as well as $t$. However as a first demonstration of this learning scheme we make a simpler choice of using a piecewise constant function (see Fig. \ref{FIGLayNetDiag}b) that is easy to conceptualize and produces simple learning rules for the weight parameter updates
\begin{equation}
\label{EQwi}
w_j(t)=\sum_{k=1}^K \omega_k\left[ S(t+(k-1)\tau)-S(t+k\tau) \right ],
\end{equation}
where $S$ is the Heaviside step function, $\tau$ defines the width of each of the $K$ pieces of $w_j$, and the $\omega_k$ are modified by learning.
We simplify this notation to use
\begin{equation}
w_{jk}= \omega_k\left[ S(t+(k-1)\tau)-S(t+k\tau) \right],
\end{equation}
where $w_{jk}$ are effectively the time delayed weights in the network. For time delays greater than $\tau K$, the connectivity weight is zero, meaning that only activity histories $H$ of length $\tau K$ are used in calculating $Q_o$.

Assuming $H$ is composed of spikes modeled as delta functions, Eq. (\ref{EQQdef}) becomes a sum of weights multiplied by the numbers of spikes
\begin{equation}
Q_o(t)=\sum w_{jk}h_{jt'},
\end{equation}
The following simple and fast weight update rules satisfy Eqs. (\ref{LRD}) and (\ref{LRU}), though other choices are possible. A weight update rule $d$ that implements $D$ when $H$ occurs is
\begin{equation}
\label{LRd}
d(w)=w-\epsilon h Q,
\end{equation}
and a corresponding weight update rule $u$ that implements $U$ when supervision spikes $o$ occur is
\begin{equation}
\label{LRu}
u(w) =w +\epsilon h o,
\end{equation}
where $\epsilon$ is a hyperparameter of the learning rules and should be chosen to be appropriately small. These learning rules cause $Q_o$ to fluctuate within a small range of $Q(o|H)$ and it may be useful to change $\epsilon$ with time to allow a initial period of fast convergence from the initialization point, and then a reduced fluctuation error once $Q_o \approx Q(o|H)$. Again, these rules are not specific to the ReLUs that we demonstrate with, these neurons could be replaced with sigmoid units, for example, without changing these weight update rules.

We use the same weight update rules for learning to predict the activity of the input layer from the activity of the hidden layers, where during learning the prediction neurons are supervised by the future input, see Fig. \ref{FIGSpikeSchem}.

\subsection{DVS MNIST Event Based Dataset}
To demonstrate this learning scheme we use a dataset collected using a Dynamic Vision Sensor (DVS) \citep{IEEESSC:2013}. The DVS is a type of video camera that collects event data, unlike conventional video cameras that collect frame data. In the camera an event is triggered by the light intensity impinging upon a pixel changing above a threshold amount. Upon such an event, the camera outputs the pixel coordinates, a timestamp (in $\mu s$) and the polarity of the change in intensity.

The MNIST database \citep{MNIST} has been used extensively in the development of deep learning \citep{bengio2014deep}. With the view of linking this work to previous work in deep learning, we demonstrate this learning scheme using a DVS version of the MNIST database \citep{DVSMNIST, MNIST} in which the handwritten digits are displayed and moved on an LCD screen that is being recorded by a DVS camera. In this dataset the light intensity changes collected by the DVS camera are primarily edges of the moving MNIST digits; however, in general the camera also captures other scene changes such as changes in illumination. The resulting dataset is noisy. Viewing the recorded data reveals that the edges are often blurred, and the number of events captured is not uniform across a digit's edges. The dataset also appears to contain some events that are not related to the movement of the MNIST digit on the LCD screen; however, these events are relatively few in number. The dataset contains recordings of 1000 handwritten digits for each integer from 0 to 9. We use the first 900 entries for training and the last 100 entries for testing. The DVS's $128\times 128$ array of pixels is cropped down to $23\times 23$ pixels with each of these pixels mapped onto two input neurons, one for each polarity of light intensity change. The input training sequence was formed from a random selection of the individual MNIST digit sequences each separated by 15 timesteps or 75 ms of no input. Each individual MNIST event sequence has a duration of about 77 timesteps or about 2.3 s.

Each pair of neurons have five $\omega_k$ parameters encoding weights for connection delays $k\tau$ of width 30 ms corresponding to the network's execution timesteps of 30 ms. In this demonstration we predict 15 timesteps or 450ms into the future. An additional ten output neurons are used to classify the current input as a digit from zero to nine. All connection weights $\omega$  between layers were initialized to small random values to the range $[0, \epsilon ]$ where $\epsilon$ was initially set to $1\times 10^{-5}$. The connection weights for the prediction and classification neurons were all initialized to zero and used an initial value of $\epsilon$ of $2.5\times 10^{-6}$, corresponding to the $\epsilon$ value for the between layer connections divided by the number of layers, since the prediction and classification neurons connect to all layers. The connections between hidden layers were trained one layer at a time for one pass through the training dataset, corresponding to $8.8 \times 10^5$ timesteps. After each pass through the hidden layers $\epsilon$ was decreased by half for all connections and training was repeated, beginning at the first hidden layer. Note that the initial value, decay and decay period for $\epsilon$ are not heavily optimized. The prediction and classification weights from all hidden layers were trained at every timestep. To demonstrate that many ideas used in deep learning are directly transferable to a spiking neural network that learns using this scheme, during training we use 50\% dropout \citep{JMLR2014} for each hidden layer.

\begin{figure}
\includegraphics[width=1\textwidth]{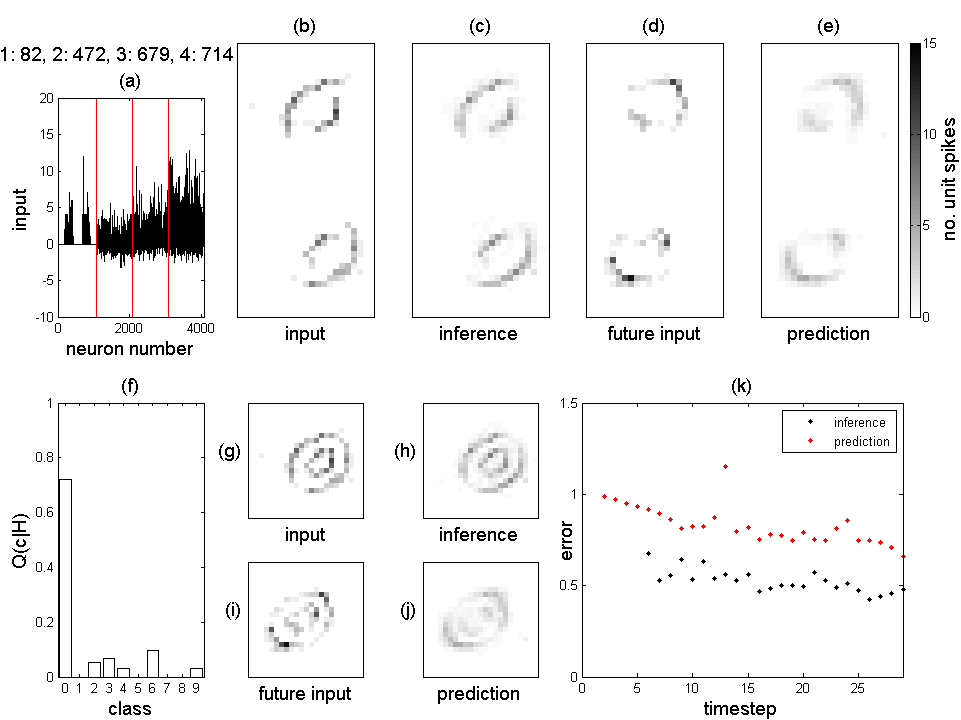}
\caption{\label{FIGDemo} A demonstration of the feedforward network described in the text applied to the DVS MNIST dataset. (a) The present input to each neuron. Vertical red lines divide layers. Neurons 1 to 1058 are input neurons, thereafter each 1000 neurons form successive hidden layers. The number of presently active neurons in each layer are indicated at the top of this frame. (b) The activity of the DVS input delayed by 5 timesteps (corresponding to the maximum connection delay). (c) The inferred activity of the input $Q(i|H)$ from the recent activity of first hidden layer. (d) The activity of the DVS input 15 timesteps into the future. (e) The network's prediction $Q(p|H)$ of the activity of the input 15 timesteps into the future. (f) Classification of the present input $Q(c|H)$. (g)-(j) As for (b)-(e) with polarity removed by summing the activity of both polarities. (k) Sum of squared errors normalized by the sum of squares of the data at each timestep for the inferences and predictions in (c) and (e). An additional file is available to  view this figure as a movie.}
\end{figure}

\begin{figure}
\includegraphics[width=1\textwidth]{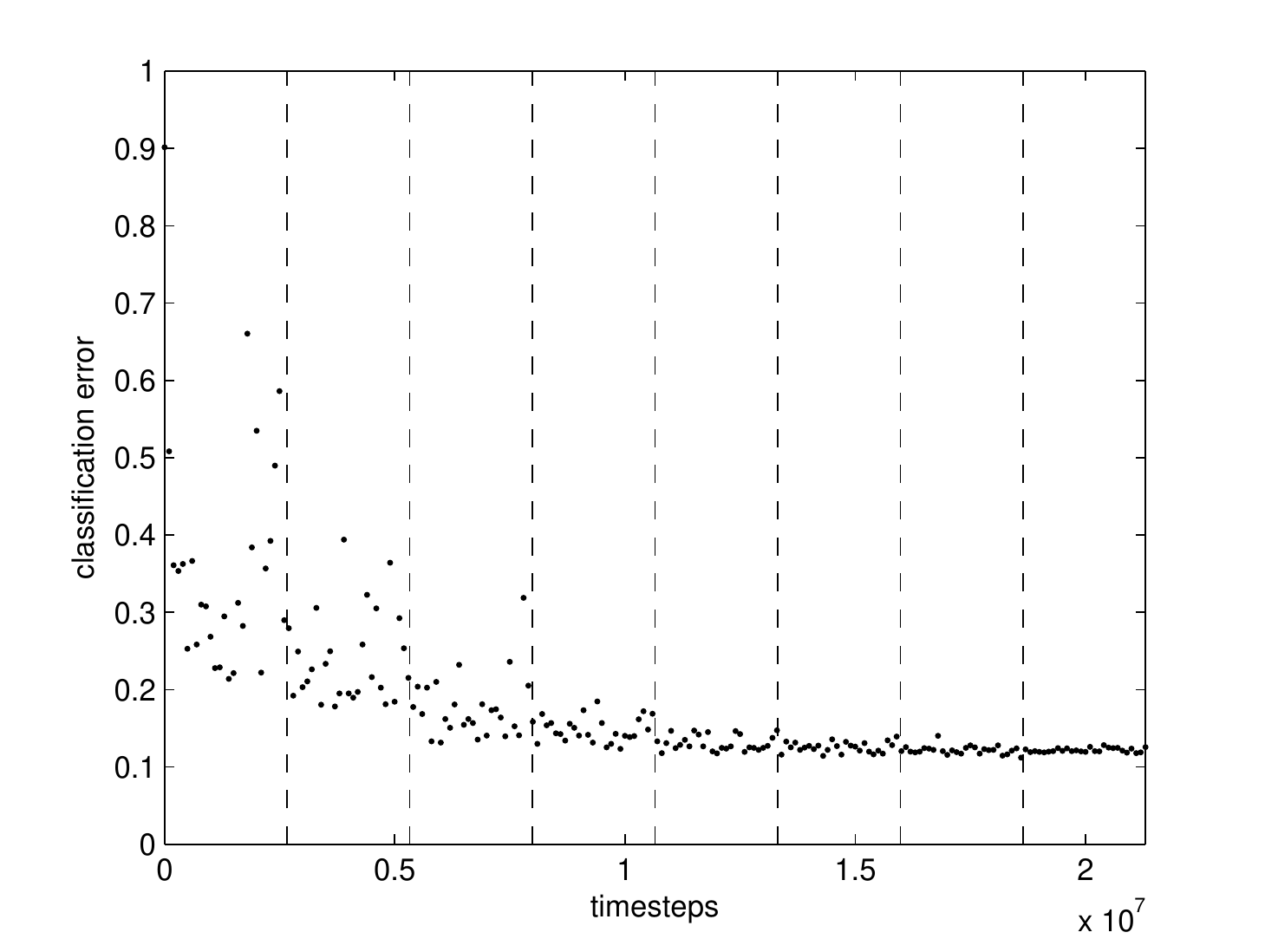}
\caption {\label{FIGClassPerf} Network classification error on the test set vs training time. Connections between layers are trained in a sequence from lowest to highest, vertical dashed lines indicate the end of each pass through the network, and the points at which $\epsilon$ is halved. At the end of training the error on the test set is 12.59\% }
\end{figure}

The operation of the trained network is demonstrated in Fig. \ref{FIGDemo}.  The hidden layers are very active since in this demonstration the neurons have a threshold fixed at zero. Including a learnable threshold would produce a more sparse representation whilst also reducing the required cpu time as the network's operation and learning are both dependent on the number of events that occur. The inference of the noisy input is significantly better than the prediction since the inference involves a memory of the input whereas the prediction does not. However, a smoothed version of the future input is usually identifiable in the prediction. The inference and predictions are often poor when the digit changes direction as the edges at these points are weak and the data are particularly noisy. The classification output correctly classifies the input digit 87.41\% of the time. The classification error as a function of training time is shown in Fig. \ref{FIGClassPerf}.

\begin{figure}
\includegraphics[width=1\textwidth]{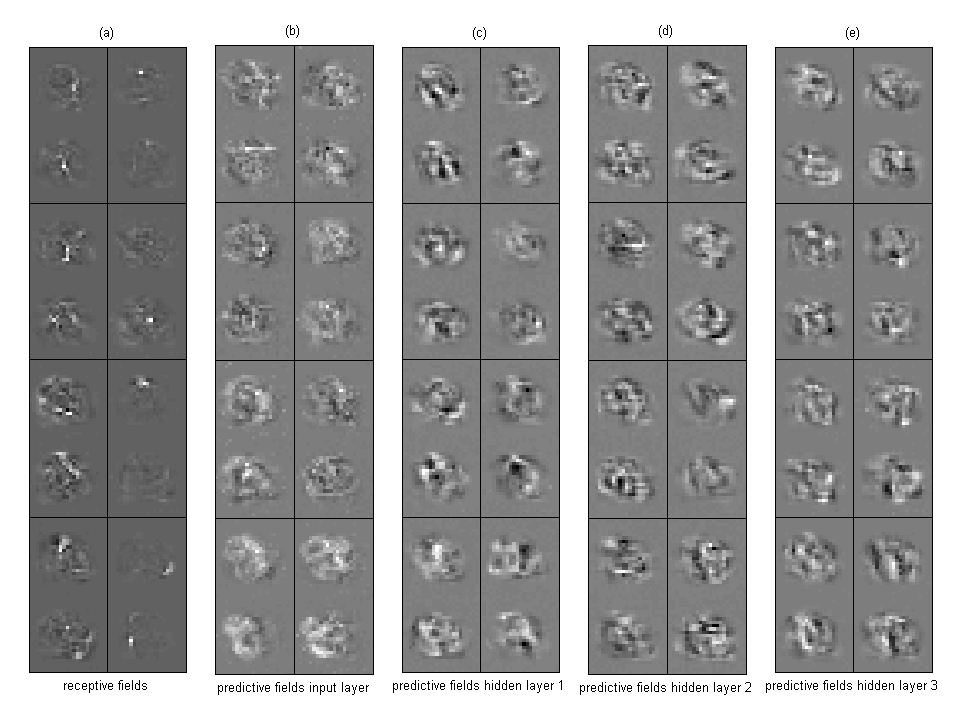}
\caption { \label{FigReceptiveFields} (a) Example receptive fields from eight neurons in the first hidden layer. (b)-(e) Example predictive fields for neurons in the input layer and hidden layers one to three respectively. In all frames each pixel is the sum of all temporal connection weights $\omega$ for that pixel. All fields have been normalized to have equal maximums.}
\end{figure}

Figure \ref{FigReceptiveFields} shows receptive fields of neurons from the first hidden layer and predictive fields of neurons from all layers. Both positive (excitatory) and negative (inhibitory) weights are learnt. Initially all neurons are active and the small random weight vectors converge toward a time averaged input vector. Upon converging toward the time averaged input, the weight vectors are nearly identical; however, differences due to the small random initialization breaks their symmetry and the weights of different neurons begin to diverge toward other more specific features of the input. This process continues as these features themselves are further split into other even more specific features. After learning is stopped, some of the receptive fields are tuned toward responding to a small number of pixels, while others respond to a distributed pattern of pixels. Predictive fields tend to be composed of larger patches of the sensory field indicating that the encoding of the prediction is distributed across many neurons. Without dropout, denoising autoencoding or another regularization method, the connectivity between hidden layers forms an identity mapping, with each neuron connecting only to a single neuron in the previous layer.

\section {Summary}

This paper introduces an event based learning scheme for neural networks. The scheme does not depend on the specific form of the neuronal dynamics or activation function, and while this paper focuses on training spiking neural networks, this scheme may also be used to train traditional artificial neural networks, especially those that involve discontinuous activation functions that defeat gradient descent methods. The scheme may also be applied to networks of neurons containing biologically inspired dynamics. Future work in this direction may inform theories of dynamics and learning in the brain. The broad applicability of this learning scheme provides an avenue to directly apply ideas from both deep learning and computational neuroscience and thus strengthen and inform the theoretical progress in both fields.





 \bibliographystyle{elsarticle-num} 
\bibliography{Learning Theory Arxiv}

\end{document}